\pdfoutput=1
\documentclass[conference]{IEEEtran}
\IEEEoverridecommandlockouts
\usepackage{hyperref}
\usepackage{amsmath,amssymb,amsthm}
\usepackage{stmaryrd}
\usepackage{algorithmic}
\usepackage{graphicx}
\usepackage{epstopdf} 
\usepackage{subcaption}
\usepackage{textcomp}

\newcommand{\appropto}{\mathrel{\vcenter{
  \offinterlineskip\halign{\hfil$##$\cr
    \propto\cr\noalign{\kern2pt}\sim\cr\noalign{\kern-2pt}}}}}

\def\BibTeX{{\rm B\kern-.05em{\sc i\kern-.025em b}\kern-.08em
    T\kern-.1667em\lower.7ex\hbox{E}\kern-.125emX}}
\begin{document}

\title{Curve Fitting from Probabilistic Emissions and Applications to Dynamic Item Response Theory
}

\author{\IEEEauthorblockN{Ajay Shanker Tripathi}
\IEEEauthorblockA{\textit{Department of Electrical Engineering}\\
\textit{Stanford University}}
\and
\IEEEauthorblockN{Benjamin W. Domingue}
\IEEEauthorblockA{\textit{Graduate School of Education}\\
\textit{Stanford University}}
}

\maketitle

\begin{abstract}
  Item response theory (IRT) models are widely used in psychometrics and
  educational measurement, being deployed in many high stakes tests such as
  the GRE aptitude test. IRT has largely focused on estimation of a single
  latent trait (e.g. ability) that remains static through the collection of
  item responses. However, in contemporary settings where item responses
  are being continuously collected, such as Massive Open Online Courses
  (MOOCs), interest will naturally be on the dynamics of ability, thus
  complicating usage of traditional IRT models. We propose DynAEsti, an
  augmentation of the traditional IRT Expectation Maximization algorithm
  that allows ability to be a continuously varying curve over time. In the
  process, we develop CurvFiFE, a novel non-parametric continuous-time
  technique that handles the curve-fitting/regression problem extended to
  address more general probabilistic emissions (as opposed to simply noisy
  data points).  Furthermore, to accomplish this, we develop a novel
  technique called grafting, which can successfully approximate
  distributions represented by graphical models when other popular
  techniques like Loopy Belief Propogation (LBP) and Variational Inference
  (VI) fail. The performance of DynAEsti is evaluated through simulation,
  where we achieve results comparable to the optimal of what is observed in
  the static ability scenario. Finally, DynAEsti is applied to a
  longitudinal performance dataset (80-years of competitive golf at the
  18-hole Masters Tournament) to demonstrate its ability to recover key
  properties of human performance and the heterogeneous characteristics of
  the different holes. Python code for CurvFiFE and DynAEsti is publicly
  available at github.com/chausies/DynAEstiAndCurvFiFE. This is the full
  version of our ICDM 2019 paper.
\end{abstract}

\section{Introduction}\label{intro}

Traditional Item Response Theory (IRT) provides powerful tools for
simultaneous analysis of both the latent traits of examinees and the
characteristics of test items. As such, IRT models are quite popular. For
example, many high stakes educational tests, such as the Graduate Record
Examination (GRE), employ IRT. IRT is also deployed in other settings, such
as healthcare, therapy, and quality-of-life research \cite{Reeve2007,
Fries2011}. In typical usage, the IRT framework conceptualizes ability as a
\textit{static} latent trait for each of the test takers. If item responses
are collected over a drawn-out period of time (e.g., students over a
5-month semester), then ability should change dynamically across the window
of observation. Indeed, this is the key goal of educational interventions.
Students can work hard to raise their ability, or have their ability
atrophy if they don't keep studying. These fluctuations could be important
feedback to pick up on. For example, a teacher may wish to reward
improvement over time to encourage such behavior. As another example, one
may wish to try various curricula to see which are effective at
accelerating ability. Extending IRT models to cover such scenarios is a
crucial psychometric need.

Over the years, initial work has been done to extend static IRT models to
handle longitudinal and time series data \cite{Verhelst1993, Molenaar1985,
Meiser1996}. Most recently, \cite{Wang2013} proposes a treatment of this
problem, Dynamic Item Response (DIR) models, based on Dynamic Linear Models
(DLM), which are commonly used to model time series data. A crucial
drawback of this approach (and the other previous approaches) is that these
models for ability curves are highly parametric. In particular, they assume
steady growth over time. While these are perhaps apt assumptions for
specific use cases (e.g., growth in reading ability over time
\cite{Wang2013}), we have relatively limited understanding of many key
dynamics of learning at this point. Thus, it seems optimal to, if possible,
relax such parametric constraints. A more flexible approach could
potentially allow us to uncover more complex and novel features of learning
dynamics.

In this paper, we propose the DynAEsti algorithm. This is a non-parametric
approach to generalizing static IRT that allows for dynamically changing
ability curves for students, as opposed to a single static ability. The
non-parametric nature of our approach allows us to capture a wider range of
learning behaviors, such as how ability may atrophy in certain settings; we
illustrate this point empirically later in the paper.

\begin{figure}[ht]
  \centering
  \includegraphics[width=3.5in]{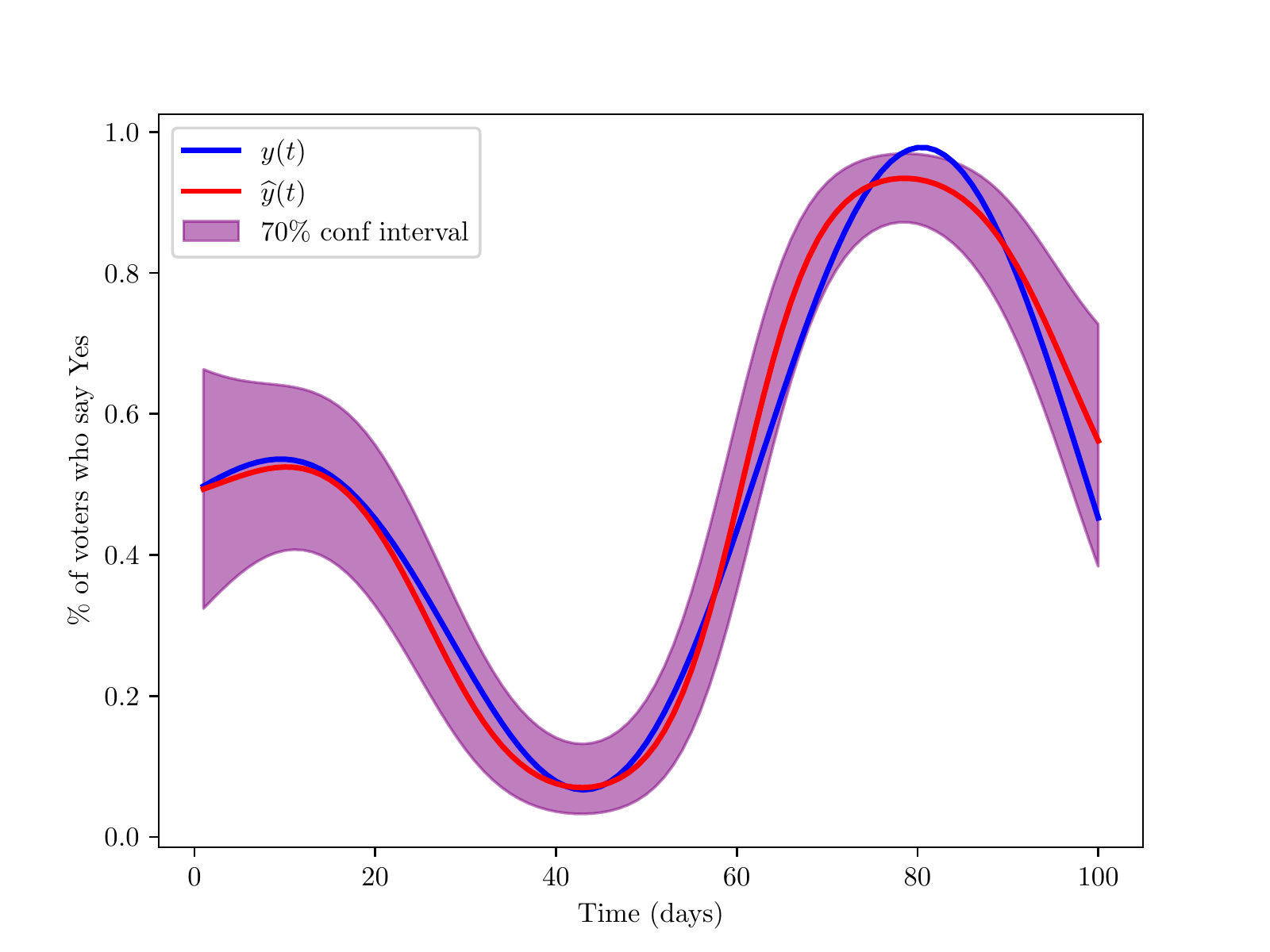}
  \caption{
    Performance of CurvFiFE demonstrated on a polling example. John Smith
    is running for president, and the percentage of people who will vote
    for him over time is estimated. $n=100$ people were polled uniformly
    spread over 100 days. $y(t)$ is the true percent of people who will
    vote for Smith. $\hat{y}(t)$ is the estimate of the curve from
    CurvFiFE. The shaded region shows the 70\% (1 std) confidence interval
    for the marginal distribution on the estimated curve at each time.
  }\label{curv_ex}
\end{figure}

Central to the DynAEsti algorithm, we address the fundamental problem of
generalizing the curve-fitting/regression problem to handle general
probabilistic emissions (as opposed to only noisy data-points). A simple
example demonstrating this problem is shown in Figure \ref{curv_ex}. John
Smith is running for president of Mars, and we wish to estimate how he is
polling in the 100 days leading up to the election. So each day, we ask a
single random Mars citizen whether they will vote for Smith. Given these
100 uniformly spaced ``emissions'', we wish to estimate the true curve
representing what percentage of people will vote John Smith over time.
Standard curve fitting techniques like spline smoothing cannot handle such
general probabilistic emissions, and can only handle noisy data-points that
were subject to symmetric Gaussian noise. In this paper, we propose
CurvFiFE, our non-parametric continuous-time solution to this problem.

Furthermore, as part of CurvFiFE, we develop \textit{grafting}, a novel
technique for approximating distributions represented as Graphical Models.
Notably, this technique proves successful in this case where popular
techniques like Loopy Belief Propogation (LBP) and Variational Inference
(VI) fail.

The paper proceeds as follows. We offer a brief introduction to IRT in
Section \ref{background}. In Section \ref{Dynamic_Ability}, we introduce
the problem of dynamic ability estimation. In Section \ref{curvfife}, we
detail the novel CurvFiFE algorithm, which provides the backbone for our
solution to the problem. Then, in Section \ref{dynaesti}, we detail
DynAEsti, our solution to the dynamic ability estimation problem.  In
Section \ref{performance}, we examine the performance of the algorithm on
simulated synthetic examples, where we achieve results comparable to the
optimal of what is observed in the static scenario. In Section \ref{golf},
we use DynAEsti to analyze a real-world dataset: 80 years of results for
hundreds of golfers at the prestigious 18-hole Masters Tournament. Finally,
in Section \ref{conclusion}, we make concluding remarks. Python code for
CurvFiFE and DynAEsti is publicly available at
github.com/chausies/DynAEstiAndCurvFiFE.

\subsection{Background on IRT}\label{background}

The prototypical setting for applications of IRT is one wherein $n$
students each respond to $m$ items. Students have static latent abilities
$\Theta = \left[\theta_1, \ldots, \theta_n\right]^T\in\mathbb{R}^n$, which
affect their item responses. $R$ is the matrix of item responses, where
$R_{ij}$ is the response of student $i$ to item $j$. For example,
$R_{ij}=1$ could indicate a correct response from student $i$ to problem
$j$.  The item responses of a student are assumed to be independent
conditioned on latent ability. Abilities are related to item responses via
$$ \Pr[R_{ij} = r | \theta_i] = F(\theta_i, \psi_j, r) $$ where $F$ is the
so-called Item Response Function (IRF), and $\psi_j$ is a vector of
parameters associated with item $j$. For example, the popular 3-parameter
logistic (3PL) IRF \cite{Birnbaum1968} for dichotomous responses $r\in\{0,
1\}$ is given
by
$$ F(\theta, a, b, c, r) = \begin{cases}
c + (1-c) \sigma\left(a (\theta - b) \right) & r=1\\
1-F(\theta, a, b, c, 1) & r=0
\end{cases}
$$
where
$$ \sigma(z) = \frac{1}{1 + \exp(-z)} $$
is the logistic function, and $\psi = (a, b, c)$ is the vector of item
parameters. For some intuition, this IRF predicts that a student with
higher ability is more likely to give a correct response. $b$ (the
difficulty) is the ``activation point'', which is around the ability needed
to start giving correct responses with decently high probability. $a$ (the
discrimination) indicates how sharp this transition is.  $c$ (the guessing
probability) gives how likely a student can give a correct response even
with ability $\theta=-\infty$.

The entire system is identified by two sets of parameters: the abilities
$\Theta$, and the item parameters $\Psi = \left[\psi_1, \ldots,
\psi_m\right]^T$. Given the $n \times m$ matrix of item responses $R$, the
log-likelihood can be written as
$$ 
  \mathcal{L}(\Theta, \Psi) = \sum_{i, j} 
    \log F(\theta_i, \psi_j, R_{ij}).
$$
This log-likelihood can be efficiently and robustly optimized for a wide
variety of IRFs using Expectation Maximization (EM). For details on this,
see \cite{Hsu2000}. In a crude approximation, the EM consists of
alternating between fixing one of $\Theta$ or $\Psi$, and optimizing
$\mathcal{L}$ w.r.t. the other; a more precise statement involves
alternating between an E-step and an M-step. In the E-step, one fixes
$\Psi$ and then finds the distribution $p(\Theta)$ of $\Theta$. Then, in
the M-step, one fixes the distribution $p(\Theta)$ and
maximizes $$\max_\Psi ~
\mathbb{E}_{p(\Theta)}\left(\mathcal{L}\left(\Theta, \Psi\right)\right)$$
to get an updated estimate for $\Psi$. Note that this expectation is taken
over the (updated) distribution $p(\Theta)$. In practice, this leads to
successful estimation of the problem parameters and the distributions on
latent abilities for students simultaneously.

One appealing feature of this formulation is that the relevant calculations
can be performed in parallel. It's straightforward to split the
log-likelihood into a sum of $n$ terms depending solely on their respective
$\theta_i$. Or it can be split into a sum of $m$ terms depending solely on
their respective $\psi_j$. So the E step can be broken into $n$ independent
optimizations and the M step can be broken into $m$ independent
optimizations. Within a step, these optimizations can be performed in
parallel.

\section{Dynamic Ability Estimation}\label{Dynamic_Ability}

We formulate the Dynamic Ability Estimation problem as follows. There are
$n$ students who respond to $m$ items. The students have latent ability
curves $\Theta (t) = \left[\theta_1 (t), \ldots, \theta_n (t)\right]^T$;
that is, student $i$ has an ability of $\theta_i (t)$ at time $t$. At time
$T_{ij}$, student $i$ responds to item $j$ to obtain a score of $R_{ij}$.
The score a student gets on an item is dependent on only their ability at
the time they respond to the item (e.g., scores are independent of prior
abilities). That is to say,
\begin{align*}
  \Pr\left[
    R_{ij} = r \vert 
    \Theta, R_{(\tilde{i}, \tilde{j}) \neq (i, j)}
  \right] &= \Pr[ R_{ij} = r | \theta_i (T_{ij}) ] \\
  &= F( \theta_i ( T_{ij} ), \psi_j, r )
\end{align*}
where $F$ is some IRF with time-independent item parameters $\psi_j$. 

To summarize, there are two data components of the dynamic ability
estimation problem: $R$, the $n \times m$ matrix of item responses, and $T
\in \mathbb{R}^{n \times m}$, the matrix of response times for each of the
students to each of the items. There are also two sets of unobserved
parameters. The abilities for each student over time are captured by
$\Theta(t)$, and $\Psi = \left[\psi_1, \ldots, \psi_m\right]^T$ contains
the parameters for all of the items. The problem is to estimate $\Theta$
and $\Psi$ given $R$ and $T$. In the next two sections, we first detail the
CurvFiFE algorithm, which is the backbone for our solution, enabling us to
estimate $\theta_i$. We then detail our solution to the overall problem:
DynAEsti.

\subsection{CurvFiFE}\label{curvfife}

CurvFiFE (pronounced ``covfefe'') is short for \textbf{Curv}e
\textbf{Fi}tting \textbf{F}rom \textbf{E}missions. It is a novel
mathematical tool we developed to solve the general problem of fitting a
curve when given ``emissions'', which is a generalization of the regression
problem. Normally, one is given many noisy data points, and tries to fit a
smooth curve that's ``close'' to them in some sense. This is a largely
solved problem, with smoothing splines \cite{Reinsch1967} being a standout
solution. We focus on a challenging generalization. Instead of being given
observations that are points on the curve plus noise, we instead observe
``emissions'': values related to a point on the curve at a given time
through a general conditional probability distribution.

More concretely, let us say there is a curve $y(t)$ which we wish to
estimate. We observe triplets $(t_i, e_i, f_i)$, $i\in\{1,\ldots, n\}$.
These triplets identify the time $t_i$ at which an emission is observed as
well as the emission $e_i$; the emission relates to $y\left(t_i\right)$,
the curve's value at the given time, through the \textit{emission
distribution} $f_i$,
$$ f_i(y(t_i))=\Pr\left[\text{observing $e_i$ at time $t_i$} | y\left(t_i\right)\right].$$
From these emissions (and their distributions), we aim to estimate, or even
find a distribution on, the curve $y$. That is, given any set of times
$T=\left[\tau_1,\ldots,\tau_m\right]^T$, one would like to estimate the
joint distribution of $y(T) = \left[y(\tau_1), \ldots, y(\tau_m)\right]^T$,
the values of the curve at these times. This joint distribution on the
curve at any set of times will be referred to as as the \textit{curve
distribution} $\mathcal{P}(y(T))$.

Returning to our polling example, Smith has a percentage of people who will
vote for him $y(t)\in[0,1]$ that is changing over time. At different times
$t_i$, a random person is polled by asking if they will vote for him,
getting a response (i.e., an emission)
$e_i\in\{\textit{Yes},\textit{No}\}$. In this simple case, the emission
distribution is a Bernoulli with parameter $y(t_i)$. In particular,
assuming they said $\textit{Yes}$, the corresponding emission distribution
will be 
$$f_i\left(y(t_i)\right) = \textit{Bern}(y(t_i); \textit{Yes}) = y(t_i)$$
If they said \textit{No}, then the emission distribution would be 
$$f_i\left(y(t_i)\right) = \textit{Bern}(y(t_i); \textit{No}) = 1-y(t_i)$$

Traditional regression methods do not work here. For example, spline
interpolation assumes that the noisy data points (which are indeed
emissions) all come from symmetric Gaussian emission distributions, which
is not necessarily always the case, as in the previous simple polling
example where emissions were from non-symmetric Bernoulli
distributions\footnote{Technically, these are Beta distributions, but for
simplicity's sake, we use the more recognizable Bernoulli terminology.}.

To solve this problem, we propose CurvFiFE. It is non-parametric, efficient
(with its most costly operation being a constant number of
matrix-multiplies), and has the very useful property that the resulting
curve distribution $\mathcal{P}(y(T))$ it estimates is just a carefully
chosen multivariate Gaussian distribution and thus easy to operate upon.

To summarize, given a set of emission triplets $(t_1, e_1, f_1), \ldots,
(t_n, e_n, f_n)$, CurvFiFE learns a curve distribution which will tell you
$$\mathcal{P}\left(y(T)\right) = \mathcal{N}(\mu, \Sigma; y(T))$$
where $\mu=\left[\mu_1,\ldots,\mu_m\right]^T$ is the estimated means (e.g.
$\mathbb{E}(y(\tau_1))=\mu_1$) and $\Sigma$ is the covariance matrix.

First, we assume that the range for the curve $y$ is $(-\infty, \infty)$,
and that the marginal distribution of $y(t)$ is $\mathcal{N}(0, 1)$ (the
standard Gaussian) for all $t$. If this is not the case, then one need only
apply a transform function $Q$ to the original curve so that it has the
desired properties. For example, in the polling case where $y(t)$'s range
is $[0,1]$ and perhaps had a Uniform prior, one could apply the probit
transform $\Phi^{-1}(y(t))$ to get the curve to obey the assumed
properties. After running CurvFiFE, one could apply the inverse transform
$Q^{-1}$ to analyze the curve in its original space.

Broadly speaking, CurvFiFE finds the curve distribution through three steps:
\begin{enumerate}
  \item\label{C1} Assume that all curves come from a reasonable prior
    distribution.
  \item\label{C2} Given the prior distribution on the curve and some
    evidence (emissions), there is a theoretical (but intractable) form for
    the posterior distribution.
  \item\label{C3} Approximate the posterior distribution as something
    tractable.
\end{enumerate}
We emphasize that this approach is firmly rooted in probability theory, as
opposed to relying on ad hoc heuristics. We will now detail and justify
these steps. 

In \hyperref[C1]{\textbf{Step 1}}, the goal is to find a prior distribution
on curves with the following properties:
\begin{enumerate}
  \item Encodes properties of curves we would like to fit (e.g. they're smooth),
  \item Can capture a wide variety of behaviors (e.g. curves that increase,
    decrease, oscillate, etc.),
  \item Is tractable and computationally convenient.
\end{enumerate}
A flexible prior with these properties is the Gaussian Process prior
\cite{Rasmussen2004}. Under this prior, the probability that a curve takes
on values $y_1, \ldots, y_m$ at times $\tau_1, \ldots, \tau_m$ is
distributed as a multivariate Gaussian $\mathcal{N}(0, \Sigma_K)$, where
the mean is $0$ and the covariance between any two points is 
$$\llbracket\Sigma_K\rrbracket_{ij} = Cov(y_i, y_j) = K(|\tau_i - \tau_j|),$$
where $K(\Delta \tau)$ is known as the \textit{covariance function}. For
our purposes, it will be the Radial Basis Function (RBF)
\begin{equation}\label{rbf}
  K(\Delta \tau) = 
  S^2 \exp \left( -\frac{1}{2} \left(\frac{\Delta \tau}{h}\right)^2\right).
\end{equation}

\begin{figure}[ht]
  \centering
  \includegraphics[width=3.5in]{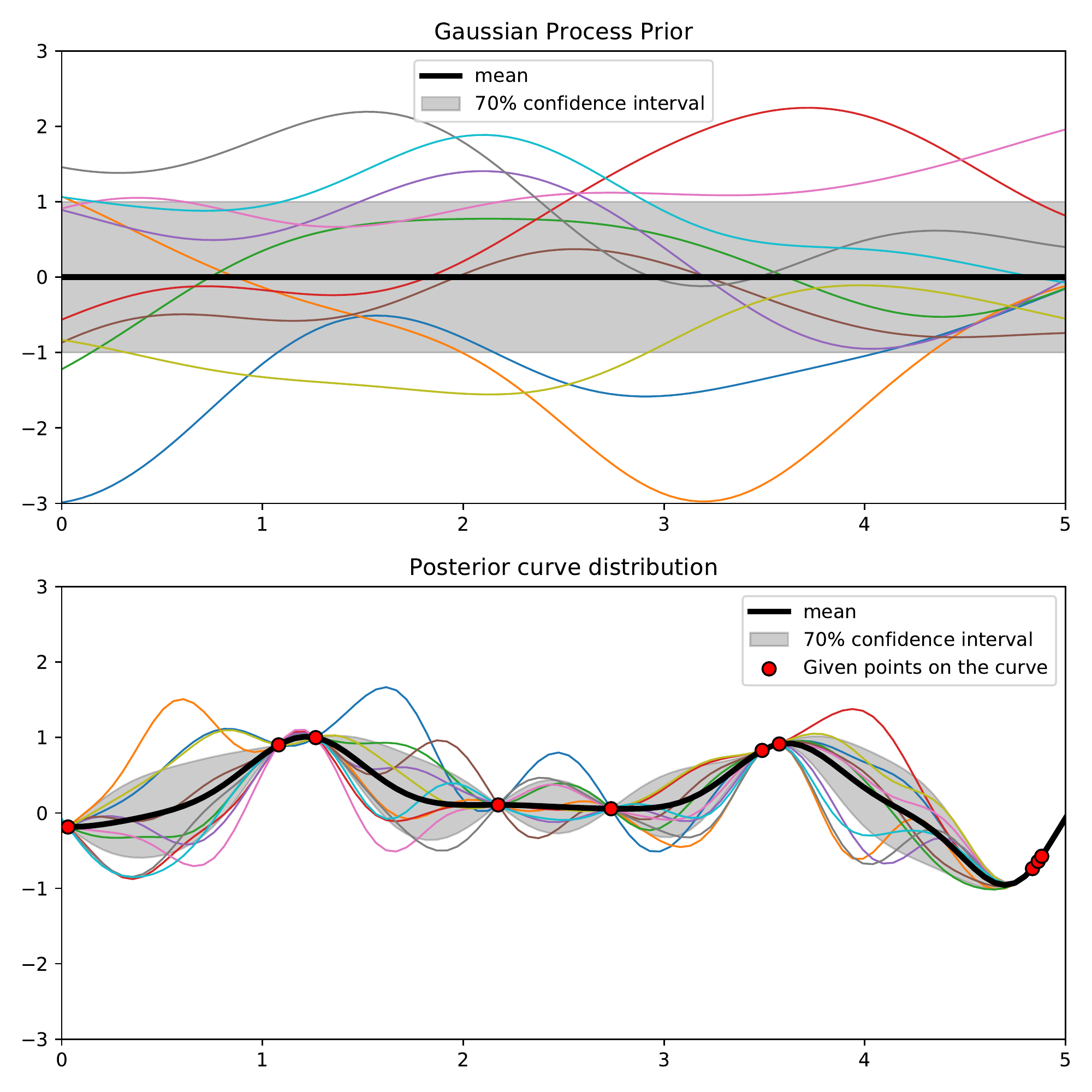}
  \caption{
    Demonstration of the Gaussian Process prior. In the first plot, sample
    curves from the prior are plotted, along with the average curve in bold
    black, and the 1 standard deviation confidence interval shaded in. In
    the second plot, the curve is fixed to go through the red points,
    yielding a posterior distribution on the curve. Again, samples, the
    mean, and the 1 std. confidence interval are plotted for this posterior
    distribution.
  }\label{krige_example}
\end{figure}

The intuition is that two temporally proximate points will have a
correlation near 1 whereas points further apart will have weaker
correlations. This powerful model for continuous-time regression also
ensures that higher-order derivatives are smooth. We demonstrate the power
of this model in Figure \ref{krige_example}. The bandwidth $h$ controls how
far apart in time two points must be before they start becoming less
correlated; for a small $h$, curves can fluctuate rapidly in time. $S$
controls the magnitude of the curve's fluctuations.  For our purposes, we
set $S=1$, since we assumed that points on the curve have a prior marginal
distribution of $\mathcal{N}(0,1)$. 

In \hyperref[C2]{\textbf{Step 2}}, we consider evidence regarding the curve
in the form of emissions, so we can consider the posterior distribution on
the curve. Here, we make the note that, for now, we will only consider the
curve points at the times of the emissions, $\left[y(t_1), \ldots,
y(t_n)\right]^T = \left[y_1, \ldots, y_n\right]^T = Y$ (i.e., we do not
consider $y(t)$ for $t$ at times where we do not observe emissions). The
posterior distribution on these points, according to Bayes Rule, will be
$$
  p(Y | e_1, \ldots, e_n) \propto 
    \mathcal{N}(0, \Sigma_K; Y)\cdot \prod_{i=1}^n f_i(y_i).
$$
In general, this is intractable to deal with. Even finding the posterior
marginal distribution of a single $y_i$ is intractable. The one standout
exception is if the emission distributions are Gaussian, because, in that
case, you're multiplying together a bunch of Gaussian factors, which will
result in a Gaussian distribution that one could easily compute and do
inference on.

\begin{figure}[ht]
  \centering
  \includegraphics[width=3in]{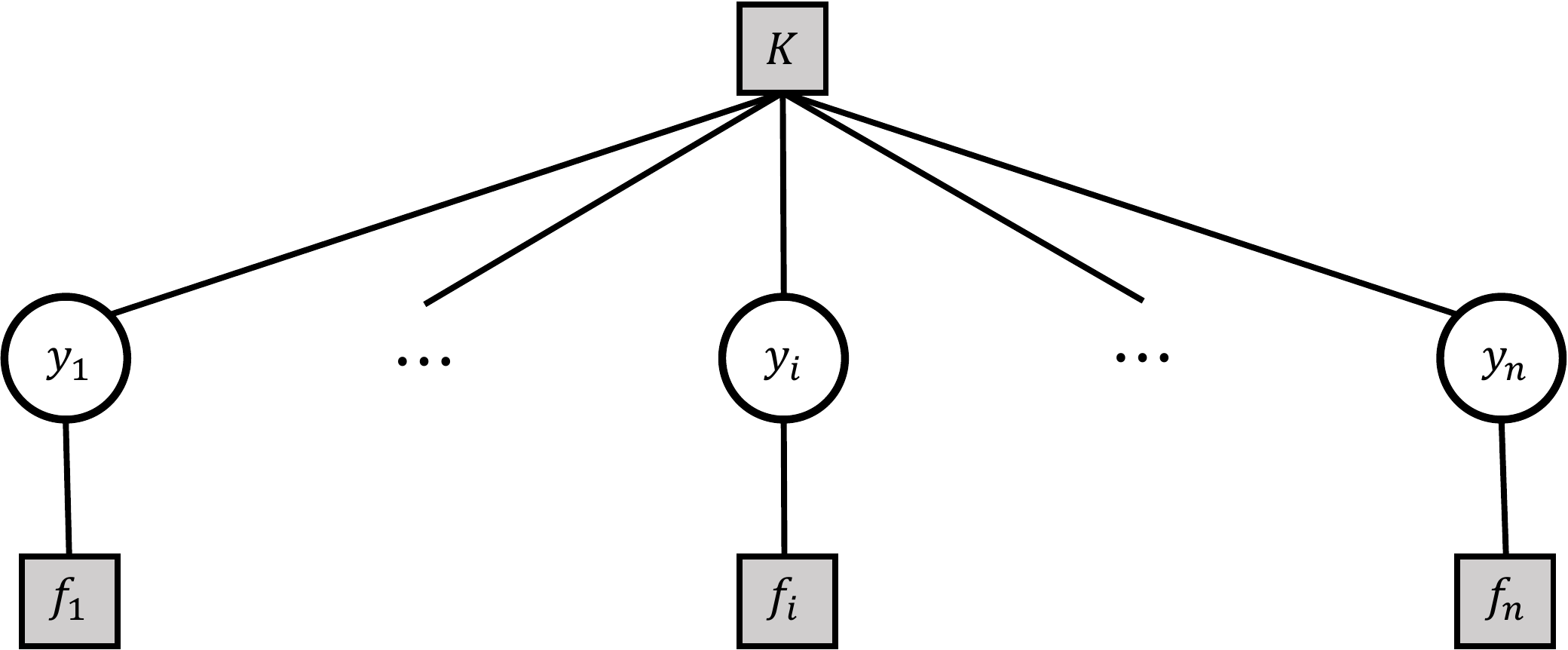}
  \caption{
    The factor graph corresponding to the curve $y$ at times $\tau_1,
    \ldots, \tau_n$. Each circle represents a variable, and each square
    represents a factor. Note that the $K$ factor represents the Gaussian
    Process prior $\mathcal{N}(0, \Sigma_K)$.
  }\label{fac_graph}
\end{figure}

In \hyperref[C3]{\textbf{Step 3}}, we approximate the posterior
distribution with a tractable alternative. We take a step back, and
consider the problem as a Graphical Model; specifically a factor graph,
which can be seen in Figure \ref{fac_graph}.

There are many established methods for trying to deal with factor graphs.
However, all of these traditional methods and their variations completely
fail for one reason or another. Just to address some of these:
\begin{enumerate}
  \item Loopy Belief Propagation (LBP) is a popular method for estimating
    the posterior marginals for the variables \cite{LBP}. It fails
    miserably in this case, however, because of the high degree of the $K$
    factor, making LBP tantamount to manually marginalizing out the other
    $n-1$ variables.
  \item Mean Field approximation (a type of Variational Inference [VI]) is
    a popular method for approximating an intractable distribution as a
    product of marginals \cite{Jordan1999}. However, it fails miserably in
    this case because the $K$ Gaussian factor is very nearly singular, with
    a large highest eigenvalue, and an extremely small lowest eigenvalue.
    The M-projection that VI finds mainly fits to this lowest eigenvalue,
    and so the approximation it finds is terrible.
  \item LBP with the messages restricted to be Gaussian. This fails
    because, when one of the initial messages is $f_i$, it could be
    impossible to approximate such a message with a Gaussian. In
    particular, if $f_i(y_i)=\frac{1}{1+\exp(-y_i)}$ (a logistic emission
    distribution), that will stay at 1 as $y_i\to\infty$, which no Gaussian
    could capture.
\end{enumerate}
Thus, in order to approximate this posterior distribution represented by a
factor graph, we developed a novel technique which we dubbed
\textit{grafting}.

To motivate grafting, we return again to the polling example. There, the
$f_i$'s were either $\textit{Bern}(y_i;\textit{Yes})=y_i$ or else
$\textit{Bern}(y_i;\textit{No})=1-y_i$. Imagine that the points $y_1,
\ldots, y_{10}$ are very close together in time and so they were all
essentially equal. This would mean that you could basically multiply all
their factors together since they're the same variable. Multiplying 10 of
those factors together would give one bell-shaped factor that can be
readily approximated as Gaussian. This gives the insight that, when many of
the factors next to each other are combined, they'll form a bell shape,
which could have just as easily been formed by using some appropriate
Gaussian factors instead. Grafting is a procedure by which we try to ``cut
off'' the original factors $f_i$ and replace them with suitable Gaussian
factors $g_i$ such that the overall distribution should look about the same
whether one used the $f_i$ factor or the $g_i$ factor. In a sense, we're
``cutting off'' the original factors, and ``grafting on'' Gaussian factors
in their place (similar to grafting work with tree branches).

Each $g_i$ can be represented with a mean $m_i$ and a variance $v_i$. Let
the vector of means and variances be $m$ and $v$ respectively. We find
these $g_i$'s through an iterative improvement algorithm as follows. First,
we initialize all the $g_i$'s to be standard Gaussians $\mathcal{N}(0, 1)$.
Then, in parallel, for each $g_i$, we do the following.
\begin{enumerate}
  \item Assume every other $g_{j\neq i}$ has been perfectly chosen to
    replace every $f_{j\neq i}$; we set our focus on  $g_i$.
  \item We look at the marginal posterior distribution of $y_i$
    \begin{align*}
      p(y_i) &= \int p(Y) dy_{j\neq i} \\
      &\propto f_i(y_i) \cdot \int\left(\mathcal{N}(0, \Sigma_K; Y) 
      \cdot \prod_{j\neq i} f_j(y_j) \right)dy_{j\neq i} \\
      &\appropto f_i(y_i) \cdot \int \left(\mathcal{N}(0, \Sigma_K; Y) 
      \cdot \prod_{j\neq i} g_j(y_j) \right)dy_{j\neq i} \\
      &\propto f_i(y_i) \cdot \mathcal{N}\left(\mu_m, \sigma_m^2; y_i\right) \\
      &= p_f(y_i)
    \end{align*}
    where $\mathcal{N}\left(\mu_m, \sigma_m^2\right)$ can be called the
    ``Gaussian message'' from all the other variables to $y_i$. We compute
    this message.
  \item We see that $p_f$ is just a product of two factors: a Gaussian
    message, and the original factor $f_i$. This product will be very
    nearly Gaussian. So we then choose $g_i$ such that
    $$p_g(y_i) = g_i(y_i) \cdot \mathcal{N}\left(\mu_m, \sigma_m^2; y_i\right)$$
    has the same mean and variance as $p_f(y_i)$.
\end{enumerate}
The previous steps update all the $g_i$'s in parallel. Then, one simply
runs the previous steps over and over again until the $g_i$'s converge (in
our experience, this takes roughly 15 iterations).

When the algorithm terminates, we've found Gaussian $g_i$'s that should
serve as good replacements for the original $f_i$ factors. Thus, we have
reduced the posterior to a Gaussian Process with Gaussian emission
distributions. The posterior is now computationally tractable. What's more,
with Gaussian factors, we can also deal simply with the $y(\tau)$ for all
the in-between times $\tau\neq t_1, \ldots, t_n$.

Using this approach, given emission triplets $(t_1, e_1, f_1), \ldots,
(t_n, e_n, f_n)$, we can successfully approximate the true curve
distribution via $\mathcal{P}(y(T)) = \mathcal{N}(\mu, \Sigma; y(T))$. For
implementation details, as well as the particulars of how to compute the
$\mu_m$ and $\sigma_m^2$ for the Gaussian message, and how to compute $\mu$
and $\Sigma$ given $m$ and $v$, see the \hyperref[append]{Appendix}.

Finally, we make a few notes here related to the performance of this
approach. First, the most costly operation in CurvFiFE is a constant number
of matrix multiplies to find the Gaussian messages. So the runtime of the
algorithm is $O(M(n))$, where $M(n)$ is the run-time of the matrix multiply
algorithm used. Secondly, we note that the grafting procedure makes one
major assumption which is that groups of emissions are close enough
temporally that their factors can be combined to form a bell-shaped factor.
This will usually be the case so long as one has been provided enough
meaningful emissions. However, if it's not the case, then grafting and
CurvFiFE may fail. We also note that, in the case where emission
distributions were in fact already Gaussian, then CurvFiFE converges in
exactly 1 iteration.

The last note relates to the one hyperparameter in this method: the
bandwidth $h$, which is the time-scale for the curve that decides how
rapidly or slowly the curve is allowed to change. Unless one has expert
input on what $h$ might be for the system, we recommend using $k$-fold
Cross Validation (CV). Basically, to test an $h$, divide the emissions into
$k$ sets, and for each set in turn, hold it out, run CurvFiFE using the
other sets, and note the average log-likelihood assigned to the held-out
emissions (with the averaging being done over the randomness in the curve
distribution). Performance for $h$ is benchmarked using the average over
the $k$ runs; a choice of $h$ is made by comparison of performance over
small set of possible values for $h$.

In Figure \ref{curv_ex}, one can see the performance of CurvFiFE on a
simulation of the polling example with $n=100$ uniformly placed emissions
(with emissions being a random person getting polled).

\subsection{DynAEsti}\label{dynaesti}

In this section, we present our IRT-based extension, DynAEsti (pronounced
``dynasty''), which is short for \textbf{Dyn}amic \textbf{A}bility
\textbf{Esti}mation. As in traditional IRT, we estimate parameters using
EM. When fixing the ability curves $\Theta$, estimating $\Psi$ is
relatively similar to the standard case of static abilities;  abilities at
the corresponding times are used in place of a single static ability in
updating of item parameters' estimates. More concretely, the log-likelihood
is 
\begin{align*}
  \mathcal{L}(\Psi | \Theta) &= \sum_{i, j} 
    \log F(\theta_i(T_{ij}), \psi_j, R_{ij}) \\
    &= \sum_j \left(\sum_i 
      \log F(\theta_i(T_{ij}), \psi_j, R_{ij})
      \right)
\end{align*}
However, since we don't have exact estimates for the curve $\theta_{ij} =
\theta_i(T_{ij})$, we have to average over the distribution
$p(\theta_{ij})$. And so the average likelihood function will be 
\begin{align*}
  \mathcal{L}(\Psi) &= \sum_j \left(\sum_i 
      \mathbb{E}_{p(\theta_{ij})}\left(\log F(\theta_{ij}, \psi_j, R_{ij})\right)
      \right).
\end{align*}
Each of these terms can be computed efficiently for any particular $\Psi$.
To maximize this w.r.t. $\Psi$, we use the L-BFGS-B algorithm, an efficient
quasi-Newton optimization method \cite{Byrd1995}. 

On the other side, we have fixed $\Psi$ (and therefore, fixed IRFs), and
the objective is to find smooth estimates for the ability curves $\Theta$.
Recall that each of the curves can be dealt with independently. To estimate
each of them, we use the CurvFiFE method described in Section
\ref{curvfife}. We use the IRFs as the emission distributions, and for each
ability curve $\theta_i$, we get the ability curve distribution
$\mathcal{P}(\theta_i(T_i))$, as well as the marginal probability
distributions $p(\theta_{i1}), \ldots, p(\theta_{im})$ needed for the
M-step where we estimate the parameters.

This alternation between the two steps of estimating either $\Theta$ or
$\Psi$ while leaving the other fixed (and using CurvFiFE as method for
estimating curve distributions) is the DynAEsti algorithm. Like the vanilla
IRT algorithm, this algorithm is also embarassingly parallel in exactly the
same way. Computation involved in each of the individual steps is also
efficient, using popular tools that have readily available implementations
in popular programming languages. We implement Python code for CurvFiFE and
DynAEsti, which is publicly available at
github.com/chausies/DynAEstiAndCurvFiFE.

\section{Synthetic Example Performance}\label{performance}

In this section, we apply DynAEsti to a synthetic dataset in order to
demonstrate its performance.

\subsection{Construction of Dynamic Synthetic Data}\label{construction}

For this example, we used $n=500$ students and $m=500$ items. The response
times are uniformly spread from 0 to 1. That is to say, $T_{ij}
=\frac{j-1}{m-1}$. A transformed 3PL IRF is used.
\begin{align*}
  & F(\theta, a, b, c, r) = \\ 
  & ~~~\begin{cases}
    c + (1-c)\sigma\left(a\left(\Phi^{-1}(\theta) - \Phi^{-1}(b)\right)\right) & r=1\\
    1-F(\theta, a, b, c, 1) & r=0
  \end{cases}
\end{align*}
where $\Phi^{-1}$ is the quantile function for the standard normal
distribution (a.k.a. the probit function), and $\sigma(z) =
\frac{1}{1+\exp(-z)}$ is the logistic function. Normally, $\theta$ is
assumed to have a standard normal distribution. With a probit-transform and
the modified 3PL IRF, $\theta$ can be seen as coming from a uniform prior.
This also makes it easier to demonstrate good estimation for the entire
spectrum of abilities, since abilities are confined to the range $[0, 1]$.
Parameters for the IRFs are chosen randomly: $b$ is sampled from
$\textit{Unif}[0, 1]$, $c$ is sampled from $\textit{Unif}[0, 0.2]$, and $a$
is sampled as $\exp(0.2 \cdot Z)$, where $Z \sim \mathcal{N}(0, 1)$.

We simulate ability curves as follows. First, a Latin Hypercube Sampling
(as described in \cite{Ye1998}) is done to get $n$ samples of the unit
square $(\theta_1(0), \theta_1(1)), \ldots, (\theta_n(0), \theta_n(1)) \in
\left[0, 1\right]^2$. This ensures an even sampling of starting and ending
points for the ability curves. Then, given the beginning and ending points,
the middle is filled out by sampling from a Gaussian Process with RBF
covariance function \ref{rbf} with $h=0.19$ and $S=0.6$. To learn more
details about this procedure and Gaussian Process Regression, see
\cite{Rasmussen2004}. Further note that, when necessary, the Gaussian
Process was repeatedly sampled until the curve lay between 0 and 1. 

\begin{figure}[ht]
  \centering
  \includegraphics[width=2.5in]{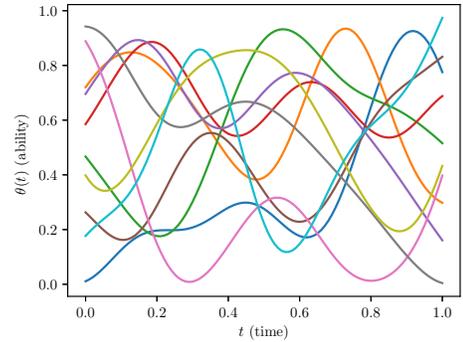}
  \caption{
    Example of the ability curves produced by the proposed Gaussian Process
    sampling procedure. For the sake of visualization, this figure
    demonstrates the curves produced when using only $n=10$.
  }\label{ex_thetas}
\end{figure}

This produces a wide range of highly varying ability curves. An example of
what these curves might look like can be seen in Figure \ref{ex_thetas}.
These curves are perhaps more ill-behaved than we would expect in practice,
but they serve as a ``stress test''; our ability to recover such curves
with fidelity would suggest that this method can handle even extremely
complex learner dynamics alongside more straightforward cases.

Lastly, the matrix of scores $R$ is filled by sampling from the true IRFs
conditioned on the true abilities at each time. With all that done, the
DynAEsti procedure is run on the $R$ and $T$ matrices to produce estimates
$\widehat{\Theta}$ and $\widehat{\Psi}$. DynAEsti is initialized with
$\widehat{a}=1$, $\widehat{b}=0.5$, and $\widehat{c}=0$ as the starting
guess for the parameters for each item. 

Note that DynAEsti produces estimates of the \textit{distribution} of each
ability curve. But, for simplicity's sake, we take the \textit{median}
ability curves to be our hard estimates and disregard the full
distributions. That is to say, for all $t \in [0, 1]$, we take
$\widehat{\theta}_i(t)$ to be the median of the estimated marginal
distribution on $\theta_i(t)$. The reason we take the median (as opposed to
the mode or mean) is because, in preliminary testing, the median estimate
performed very slightly better. Furthermore, the median is preserved by
monotonic transformations (like the probit transform we use), which may be
a desirable property.

\subsection{Parameter Recovery}\label{recovery}

\begin{figure}[ht]
  \centering
  \includegraphics[width=3.5in]{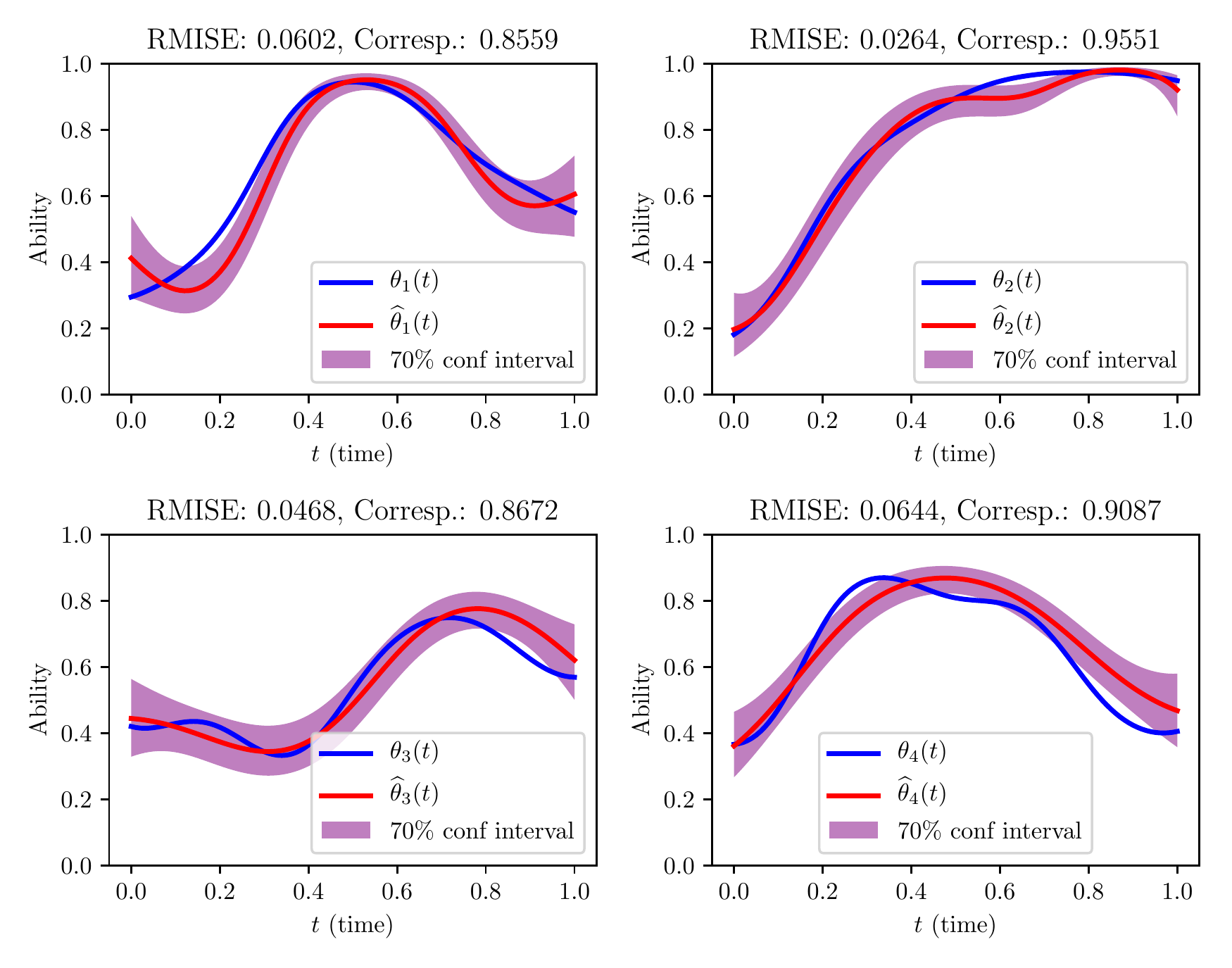}
  \caption{
    DynAEsti estimates (compared to truth) for 4 out of the 500 ability
    curves. The marginal 70\% (1 std. div.) confidence interval for those
    estimates is also shown.
  }\label{theta_hats}
\end{figure}

A few examples are provided in Figure \ref{theta_hats} to show how the
estimates for the ability curves track the true ability curves. As can be
seen, the estimates are reasonably accurate and err on the side of being
smooth. This occurs because there is not enough data to reliably fit the
higher-frequency undulations and the cross-validation scheme in CurvFiFE
automatically understands this and errs on the side of smoothness.

Here, we introduce some metrics in order to benchmark overall performance.
Note that for a given IRF, several different triplets $(a, b, c)$ can
correspond to similar looking IRFs \cite{Maris2009}, so $\left\|\psi -
\widehat{\psi}\right\|$ isn't a good measure of how far off the estimated
IRF is from the true IRF.  Instead, we use the Root Mean Integrated Squared
Error (RMISE) between the true and estimated IRFs. The RMISE between a true
function $f$ and its estimate $\widehat{f}$ is
$$RMISE(f, \widehat{f}) = \sqrt{\int_0^1 \left( f(x) - \widehat{f}(x) \right)^2 dx}.$$
This gives a measure of the difference between the IRFs, with larger errors
carrying heavier penalties.

As for the ability curves, it also makes sense to use the RMISE as a metric
for quality of estimation. We also consider an additional metric as, in
practice, one is mainly concerned with detecting fluctuations in the
ability (i.e. When does ability dip? When does it surge?). For example,
ascertaining whether learning is actually occurring and abilities aren't
remaining static could be a key objective. This motivates a metric which
measures if changes in the true ability curve are reflected well by changes
in the estimated ability curve. To capture this, we propose the
\textit{correspondence} metric (roughly speaking, the correlation between
the derivatives of the true and estimated ability curves)
$$ C(\theta, \widehat{\theta}) 
= \frac{\int_0^1 \theta^\prime(t) \cdot \widehat{\theta}^\prime(t) dt}
{
  \sqrt{\int_0^1 \left(\theta^\prime(t)\right)^2 dt} 
  \sqrt{\int_0^1 \left(\widehat{\theta}^\prime(t)\right)^2 dt}
} \in [-1, 1]$$
The correspondence between $\theta$ and $\widehat{\theta}$ is nothing but
the cosine similarity between their derivatives, and ranges from 1 (full
correspondence) to -1 (complete anti-correspondence). 

To get a sense of what is optimal, we note here that, in the hypothetical
perfectly static case (where abilities are completely static), we could use
standard IRT. In this case, we would estimate the IRFs with an RMS RMISE of
$0.039$ across all $m=500$ items, and we would estimate abilities with an
RMS error of $0.047$ across all $n=500$ students. This gives a bound on
what we could possibly achieve in the dynamic case. With these as
benchmarks, we report the  performance of DynAEsti in this simulated
example. The IRFs were estimated with an RMS RMISE of $0.049$ across all
items (compared to the optimal $0.039$). The ability curves were estimated
with an RMS RMISE of $0.091$ across all students (compared to the optimal
$0.047$). Furthermore, the average correspondence between the true and
estimated ability curves is around $0.72$, with 80\% having correspondence
over $0.6$. In the cases where problems had lower correspondence, this was
largely because the estimated curve was overly smooth while the true
ability curve undulated rapidly; such undulations couldn't be reliably
captured due insufficient emissions.

\section{Empirical Data Analysis: The Masters Golf Tournament}\label{golf}

\subsection{Background}\label{golf_background}

Having developed DynAEsti, we demonstrate its utility by analysis of data
from the Masters Golf Tournament. Considered by many to be the most
prestigious golf tournament, the Masters has been going on since 1934 (over
80 years). It is the only major golf tournament to always take place at the
exact same course: the famous Augusta National Golf Club. The course
consists of 18 holes, and golfers try to complete each hole in as few
strokes as possible. Each year, the Masters tournament sees around 100
golfers, who play the 18-hole course 4 times (rounds). Each hole has a
pre-determined number of strokes that the golfers are expected to take for
it; this is known as the hole's ``par''. Whoever has the fewest strokes
over all 4 rounds at the end wins. Some players have attended over 35 of
these tournaments. Tiger Woods, for example, has attended 21 of them.

While each stroke in golf is weighted equally in determining the tournament
winner, this could be a suboptimal scheme for estimating ability (i.e., an
unweighted sum over all strokes may be a worse predictor of future
performance than an alternative estimation scheme). With that motivation,
one may wish to identify the characteristics (e.g. the IRF) of each of the
18 holes, and how they reflect on underlying ability. However, standard IRT
doesn't allow for this. Each player's ability may vary wildly over the
decades they attend the tournament, so the assumption of a static ability
would clearly be problematic. Furthermore, ability does not simply grow,
but may also atrophy, as can be seen by comparing Tiger Woods's
record-breaking 1997 performance with some of his more recent performances.
DynAEsti is designed to address these complicating factors so as to get
high quality estimates for the IRFs of the holes, as well as the ability
curves for the hundreds of golfers.

From the official Augusta Golf Club website, we have the scorecard for each
player for each of their 4 rounds on the 18 holes for each of the years
between 1937 and 2018 (barring years like those during WWII when the
tournament wasn't properly held).

The IRF we used to model the holes was a modified Generalized Partial
Credit Model (GPCM) \cite{Masters1982}, which we call Golf IRF for
convenience.\footnote{Rather coincidentally, the Partial Credit Model was
constructed by a fellow named Geoff Masters.} For a particular hole, let
$$\omega_s(\theta) = \log\left(
  \frac
    {\Pr[\text{$s$ strokes below par}~|~\text{ability $\theta$}]}
    {\Pr[\text{par}~|~\text{ability $\theta$}]}
\right)$$
be the log odds of getting $s$ strokes below par instead of a par given one
has ability $\theta$. $s=1$ implies a ``birdie'' (one under par), and
$s=-1$ implies a ``bogey'' (one over par). Clearly, $\omega_0(\theta) = 0$.
For each stroke count $s$ (besides $s=0$), there is an $a_s\in(0, \infty)$
(discrimination) parameter and a $b_s\in(-\infty, \infty)$ (difficulty)
parameter. The log odds for $s$ is set to be
$$ \omega_s(\theta) = 
  sgn(s) \cdot \sum_{sgn(s)}^s a_s \cdot \left(\theta - b_s\right)
$$
where $sgn(s)$ is the sign ($\pm 1$) of $s$. If $r = s - sgn(s)$ is the
stroke one closer to par than $s$, then $b_s$ tells you the ability such
that scoring an $s$ and scoring an $r$ are equally likely (their odds
intersect). $a_s$ tells you how sharp the transition of odds between $r$
and $s$ is. Large $a_s$ means it's a sharp transition from the odds being
in favor of $r$ to being in favor of $s$. Small $a_s$ means a much flatter
transition. To generate the probabilities of the stroke counts, we simply
take a softmax of the log odds.

As a final note on the Golf IRF, we clipped (thresholded) the stroke counts
to only allow for stroke counts that have had over 100 instances. For
example, a triple bogey ($s=-3$) on hole \#1 has only occurred 24 times in
the history of the Masters. That's far too few samples to reliably fit
$a_{-3}$ and $b_{-3}$. Conversely, double bogeys ($s=-2$) have occurred 254
times, which is enough to fit. So all stroke counts $s<-2$ are just counted
as double bogeys on hole \#1.

\subsection{Results}\label{golf_results}

\begin{figure}[ht]
  \center
  \includegraphics[width=3.5in]{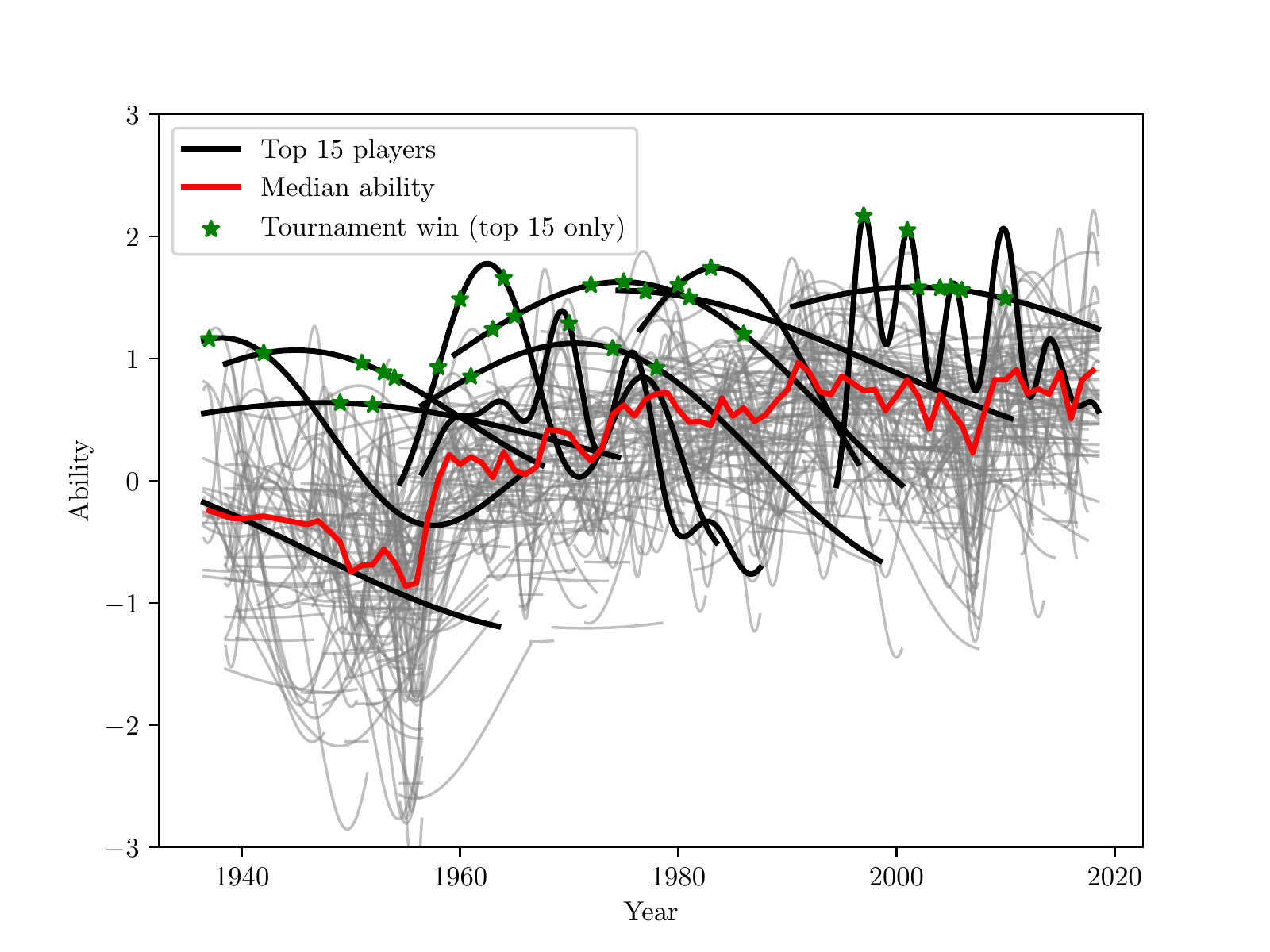}
  \caption{
    Ability curves over the years for the hundreds of Masters attendees.
    Those who only attended 1 Masters aren't included. The highlighted top
    15 golfers include Jack Nicklaus, Arnold Palmer, and Tiger Woods. The
    highest ability reached ever is Tiger Woods in 1997, during his
    record-breaking performance.
  }\label{overall_traj}
\end{figure}

\begin{figure}[ht]
  \center
  \includegraphics[width=3in]{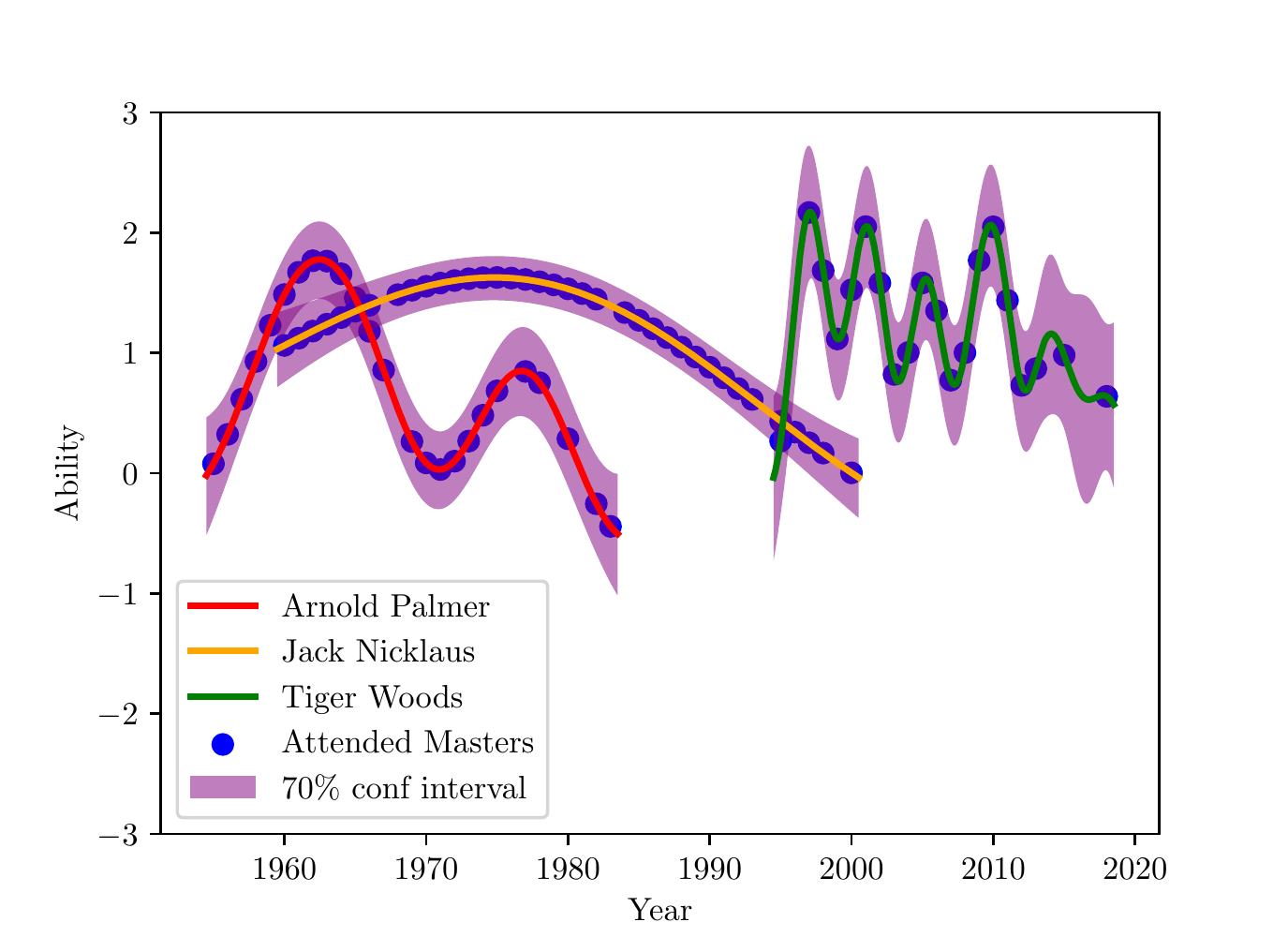}
  \caption{
    The ability curves for three legendary golfers, along with 70\% (1 std.
    div.) confidence intervals on their abilities.
  }\label{legend_traj}
\end{figure}

We use DynAEsti to estimate ability dynamics and features of the 18 holes
at Augusta. In Figure \ref{overall_traj}, we can see the trajectories of
the ability curves for the hundreds of golfers who've attended the Masters
over the years. Figure \ref{overall_traj} yields several insights. First,
over the decades, abilities have generally increased. Given developments in
golf-related technology and technique, this is quite reasonable. Second,
for golfers who have participated in many Masters, there's a common trend
that ability initially grows, peaks, and then deteriorates (with perhaps a
small ``second wind'' bump). This behavior is exemplified in Figure
\ref{legend_traj} by Jack Nicklaus and Arnold Palmer.

\begin{figure}[ht]
  \center
  \includegraphics[width=3in]{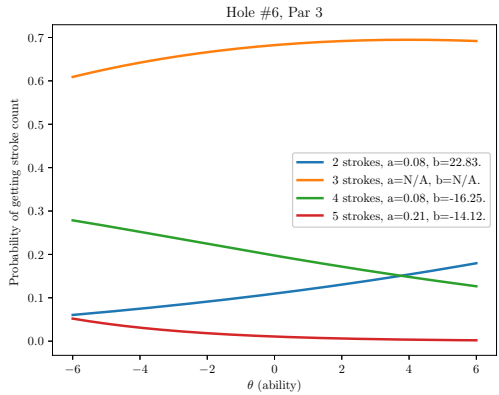}
  \caption{
    The IRF for the 6th hole of the masters, known as Juniper.
  }\label{hole_6}
\end{figure}

We now turn to the characteristics of the 18 holes. We note two main
classes of interesting behavior here. On one hand, there are holes like
hole \#6, a.k.a. Juniper, whose IRF is in Figure \ref{hole_6}. As can be
seen, performance on the hole is fairly flat with respect to ability
$\theta$. No matter where their ability lies between -2 and 2 (which is
where most of the golfers are), a golfer makes par with $\approx70\%$
chance, or gets unlucky and bogeys with $\approx20\%$ chance. Changes in
ability only change these odds slightly.  This is due to the fact the hole
has relatively low discrimination parameters $a$, and difficulty parameters
$b$ that are too large in magnitude for typical abilities to compare to. 

\begin{figure}[ht]
  \center
  \includegraphics[width=3in]{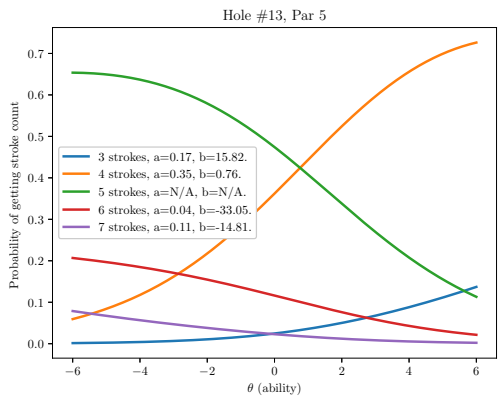}
  \caption{
    The IRF for the 13th hole of the masters, known as Azalea.
  }\label{hole_13}
\end{figure}

On the other hand, there are holes like \#13, a.k.a. Azalea, whose IRF is
in Figure \ref{hole_13}. Performance on this hole is more strongly linked
to ability. Consider the birdie ($s=1$) and par ($s=0$) curves. The ability
needed to have a strong probability of obtaining a birdie is reasonable;
moreover, the transition in odds is also fairly discriminative. Because it
better allows golfers to demonstrate their ability, hole \#13 discriminates
golfers with respect to their ability more so than does hole \#6.

With this, it's clear to see that not all strokes are created equal. For
some holes, performance is mostly chance. On very few holes, performance is
significantly affected by ability. Overall, a Masters tournament victory is
not just about having high ability, but a lot about getting pretty lucky.

\subsection{DynAEsti vs. Static IRT}\label{dyn_vs_sirt}

As a final note, we demonstrate that DynAEsti's allowance for dynamic
ability curves is necessary to accurately model this golf example. Allowing
for dynamic abilities yields far more accurate out-of-sample estimates than
does traditional static IRT (SIRT). To motivate how we judge the schemes,
consider that Arnold Palmer has participated in 25 different Masters,
``responding'' to (playing) 1800 holes in total.  We compare performance by
giving an algorithm (SIRT or DynAEsti) half of these responses to learn
about Palmer's ability, and then ask the scheme to assign a likelihood to
the remaining half.

The particulars of how we judge are as follows.
\begin{enumerate}
  \item Each scheme has to learn its own IRFs from scratch.
  \item Averaged (geometrically) over 5 different \textit{runs}:
    \begin{itemize}
      \item Randomly divide Palmer's 1800 responses into 2 halves.
      \item Hold out 1 half, let the scheme learn Palmer's ability using
        the other half, and then have the scheme assign a probability to
        observing the responses in the held-out half.
      \item Switch the roles of the two halves, and take the geometric mean
        of the result.
    \end{itemize}
\end{enumerate}

Using this, we can get the (geometric) average probability that both
DynAEsti and SIRT would assign to Palmer's responses that it didn't
observe.

As one would imagine (see Figure \ref{legend_traj}), Palmer's ability
varied significantly over the decades he's attended the Masters. SIRT does
not account for this. For example, SIRT couldn't use the fact that Palmer's
ability in the 1960's was peak, and deteriorated in later years, whereas
DynAEsti (via CurvFiFE) can pick up on that information. As a consequence,
DynAEsti handily outperformed SIRT. On (geometric) average, DynAEsti
assigned over 120 times higher likelihood to the held out responses than
SIRT did.

\section{Conclusion and Further Research}\label{conclusion}

IRT is a powerful framework for understanding item responses. In this
paper, we proposed an extension of IRT, DynAEsti, that captures dynamically
changing ability curves without relying on potentially unfounded parametric
assumptions. We showed the performance of DynAEsti is comparable to a bound
on the theoretically optimal performance. Furthermore, DynAEsti produces
estimates of the ability curves with high correspondence to the true
ability curves. As such, DynAEsti allows for high fidelity detection of
ability growth and even potential decay. Thus, it may offer useful feedback
for either the student or an administrator in digital learning environments
(e.g., MOOCs) where item responses are being continuously collected.

DynAEsti allowed us to analyze the performance of golf players at the
famous Masters tournament over the decades, where traditional static IRT
would perform poorly. We were able to detect many interesting trends in
player abilities over time as well as the characteristics of different
holes. Finally, we showed how DynAEsti drastically outperforms static IRT
in terms of its predictive power in this context.

To make this possible, we developed the CurvFiFE algorithm, which provided
an efficient and non-parametric solution to the curve-fitting/regression
problem extended to account for general probabilistic emissions. At the
heart of this was the novel grafting technique we developed, which provided
a means to approximate graphical models where standard LBP and VI
techniques failed.

\section{Appendix: CurvFiFE Implementation Details}\label{append}

Here, we will give some important implementation details for CurvFiFE that
were left out of the main body of this paper.

The first note is that, we slightly modify the covariance function used to
be 
$$
  K(\Delta \tau) = K_{\text{RBF}}(\Delta \tau) 
  + \epsilon \cdot \delta(\Delta\tau)
$$
, where $\delta(\cdot)$ is the Dirac delta function, which is equal to 1
when its argument is 0, otherwise it is 0. This is essentially like adding
a small amount onto the diagonal of the corresponding covariance matrix
$\Sigma_K$.

The reason for this modification is to help with numerical stability,
because $\Sigma_K$ is generally very nearly singular, with a high
conditional number, so inverting it often yields numerical errors on
finite-precision computers. We recommend $\epsilon=0.0001$.

Next, we give the formulae for the various calculations made by CurvFiFE.
These formulae are presented without proof, but it just involves some
simple matrix algebra, as well as the use of the Woodbury matrix identity,
to derive them.

Given the previous guess for the Gaussian factors have means $m$ and
variances $v$, the Gaussian messages with means $\eta = \left[\mu_m^{(1)},
\ldots, \mu_m^{(n)}\right]^T$ and variances $\rho =
\left[{\sigma_m^{(1)}}^2, \ldots, {\sigma_m^{(n)}}^2\right]^T$ can be
computed in parallel as follows.
\begin{align*}
  P &= \left(\Sigma_K^{-1} + Diag\left(\frac{1}{v}\right)\right)^{-1} \\
  \widehat{\mu} &= P \cdot \left(\frac{m}{v}\right) \\
  H &= P + \left(diag(P)\cdot \mathbf{1}_n^T\right) \circ P 
  / \left((v - diag(P))\mathbf{1}_n^T\right) \\
  M &= \mathbf{1}_n \left(\frac{m}{v}\right)^T 
  - Diag\left(\frac{m}{v}\right) \\
  \rho &= diag(H) \\
  \eta &= (H \circ M) \mathbf{1}_n
\end{align*}
Note that division is element-wise, and $\circ$ is element-wise
multiplication. Also, $\mathbf{1}_n$ is a length-$n$ column vector of ones.

Given the Gaussian messages, one computes the mean $\nu_i$ and variance
$\gamma_i$ of each of the marginal $p_f(y_i)$ distributions. This can be
done in a straightforward manner by discretizing the $y_i$ number line
(from say, -6 to 6), and computing the mean/variance from the discrete
approximation to $p_f(y_i) \propto f_i(y_i) \cdot
\mathcal{N}\left(\mu_m^{(i)}, {\sigma_m^{(i)}}^2 ; y_i\right)$.

Finally, given $\nu$ and $\gamma$, the following is how one computes the
$m$ and $v$ for Gaussian factors in parallel.
\begin{align*}
  v = \frac{1}{\frac{1}{\gamma} - \frac{1}{\rho}} \\
  m = v \circ \left(\frac{\nu}{\gamma} - \frac{\eta}{\rho}\right)
\end{align*}
The one thing of note here is that, sometimes, entries of $v$ may be
negative, which occurs when an $f_i$ factor is particularly unsuited to a
Gaussian approximation around the support of the $i$th Gaussian message. In
this case, one essentially throws away that emission by replacing it with a
Gaussian factor with infinite (or very large) variance $L$. In practice, we
used $L = 10^6$.

Lastly, having computed $m$ and $v$, we wish to find the distribution
$\mathcal{N}(\mu, \Sigma)$ of $y(T)$ for any times $T = [\tau_1, \ldots,
\tau_m]^T$. This is done as follows.

\begin{align*}
  \llbracket\Sigma_1\rrbracket_{ij} &= \llbracket\Sigma_K\rrbracket_{ij} 
  = K(|t_i - t_j|) \\
  \llbracket\Sigma_2\rrbracket_{ij} &= K(|\tau_i - \tau_j|) \\
  \llbracket\Sigma_{12}\rrbracket_{ij} &= K(|t_i - \tau_j|) \\
  W &= \Sigma_{12}^T \cdot \Sigma_1^{-1} \cdot 
  \left(\Sigma_1^{-1} + Diag\left(\frac{1}{v}\right)\right)^{-1} \\
  \Sigma &= \Sigma_2 + (W - \Sigma_{12}^T)\cdot\Sigma_1^{-1}\cdot\Sigma_{12} \\
  \mu &= W \cdot \left(\frac{m}{v}\right)
\end{align*}

As a final note, we implemented CurvFiFE and DynAEsti in Python 3, with all
the main operations being done with the PyTorch library, assisted in part
by Numpy and SciPy. This code for CurvFiFE and DynAEsti is publicly
available at github.com/chausies/DynAEstiAndCurvFiFE.

\bibliographystyle{ieeetr}
\bibliography{mybib}

\end{document}